\documentclass[10pt,twocolumn,letterpaper]{article}

\usepackage{wacv}              

\newcommand{\dataset}{EXAFace\xspace}

\usepackage{subscript}

\usepackage{tikz}
\usetikzlibrary{3d,arrows.meta,positioning,shapes.geometric,calc,matrix,fit}
\usepackage{tikz-3dplot}
\usepackage{tikzscale}

\usepackage{graphicx}
\usepackage{caption}
\usepackage{mathtools}
\usepackage{bm}
\usepackage{colortbl}

\usepackage[accsupp]{axessibility}  

\makeatletter
\renewcommand\paragraph{
  \@startsection{paragraph}
                {4}
                {\z@}
                {-1.0ex \@plus -0.5ex \@minus -.2ex}
                {-1em \@plus 0.2ex}
                {\normalfont\normalsize\bfseries}}
\makeatother

\definecolor{wacvblue}{rgb}{0.21,0.49,0.74}
\usepackage[pagebackref,breaklinks,colorlinks,allcolors=wacvblue]{hyperref}

\def\wacvPaperID{2902} 
\def\confName{WACV}
\def\confYear{2026}

\title{Extreme Amodal Face Detection}

\author{Changlin Song$^1$ \quad Yunzhong Hou$^1$ \quad Michael Randall Barnes$^2$ \quad Rahul Shome$^1$ \quad Dylan Campbell$^1$\\
$^1$Australian National University \quad $^2$University of Oslo\\
{\tt\small \{changlin.song, yunzhong.hou, rahul.shome, dylan.campbell\}@anu.edu.au}\\
{\tt\small michael.barnes@ifikk.uio.no}
}

\usepackage{amsthm}
\theoremstyle{definition}
\newtheorem{definition}{Definition}   
\theoremstyle{plain}

\newcommand{\xqed}[1]{%
    \leavevmode\unskip\penalty9999 \hbox{}\nobreak\hfill
    \quad\hbox{\ensuremath{#1}}}
\newcommand{\Endofdef}{\xqed{\blacksquare}}

\usepackage{xspace}
\makeatletter
\DeclareRobustCommand\onedot{\futurelet\@let@token\@onedot}
\def\@onedot{\ifx\@let@token.\else.\null\fi\xspace}

\def\eg{e.g\onedot} 
\def\ie{i.e\onedot}

\makeatother

\newcommand{\reals}{\mathbb{R}}

\newcommand{\cC}{\mathcal{C}}

\newcommand{\cS}{\mathcal{S}}

\usepackage{tabularx}
\newcolumntype{C}{>{\centering\arraybackslash}X}

\usepackage[capitalize,noabbrev]{cleveref}

\usepackage{dsfont}
\usepackage{multirow}
\usepackage{multicol}

\def\imgA{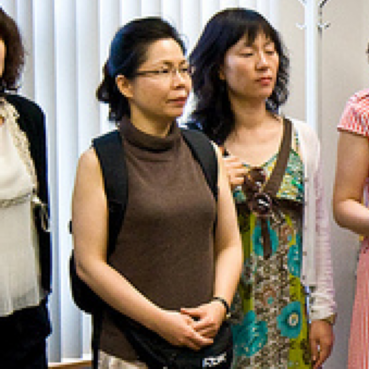}
\def\imgB{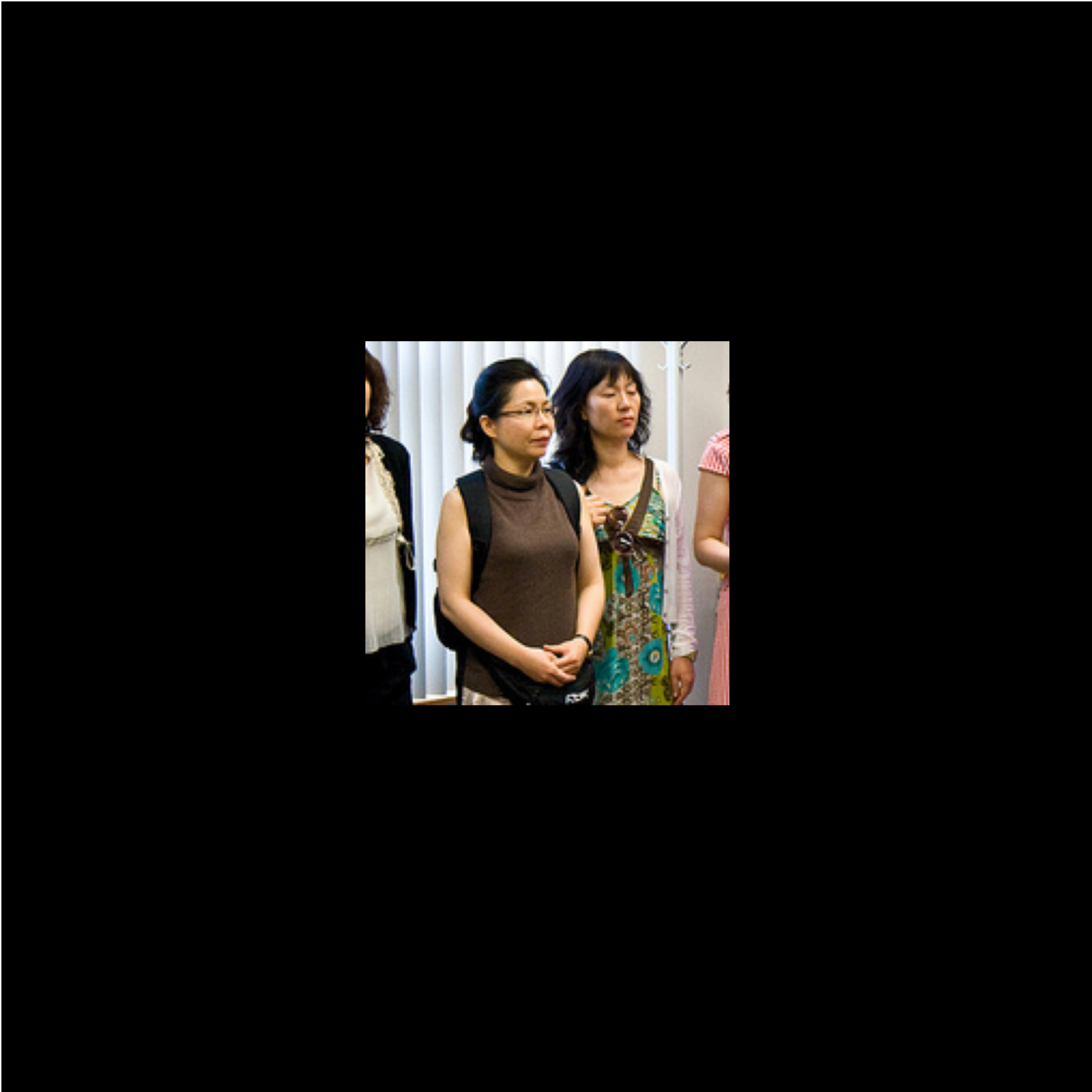}
\def\imgMap{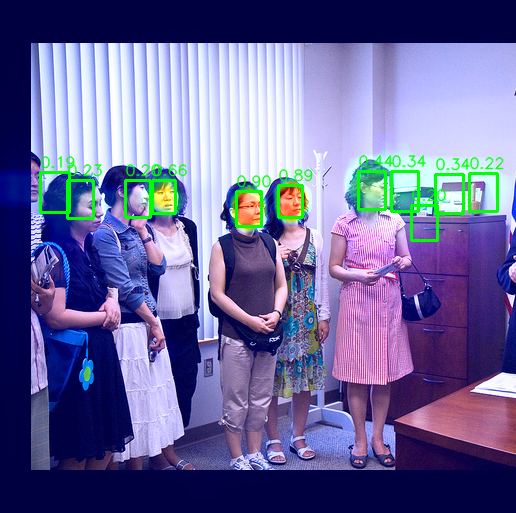} 
\def\CenterOpacity{0.5} 
\def\RingOpacity{0.5}   

\newcommand\splashfigure{%
    \centering
    \captionsetup{type=figure}
    \begin{subfigure}[]{0.31\linewidth}\centering
        \includegraphics[
        height=11em,
        trim=162pt 113pt 162pt 97pt, clip]{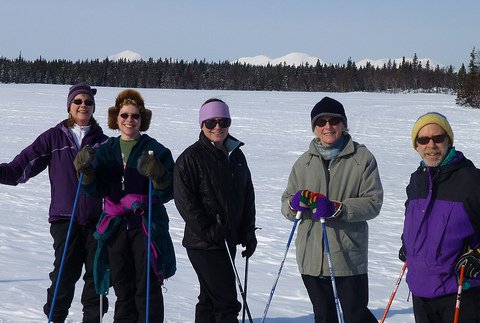}
        \caption{Input image}
        \label{fig:task_a}
    \end{subfigure}\hfill
    \begin{subfigure}[]{0.33\linewidth}\centering
        \includegraphics[
        height=11em,
        trim=0 0 0 20pt, clip]{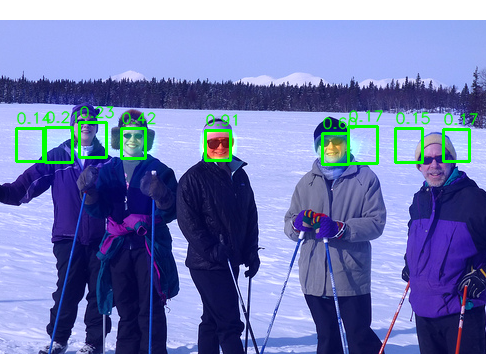}
        \caption{Extreme amodal face detection}
        \label{fig:task_b}
    \end{subfigure}\hfill
    \begin{subfigure}[]{0.33\linewidth}\centering
        \includegraphics[
        height=11em,
        trim=4pt 0 4pt 1pt, clip,
        ]{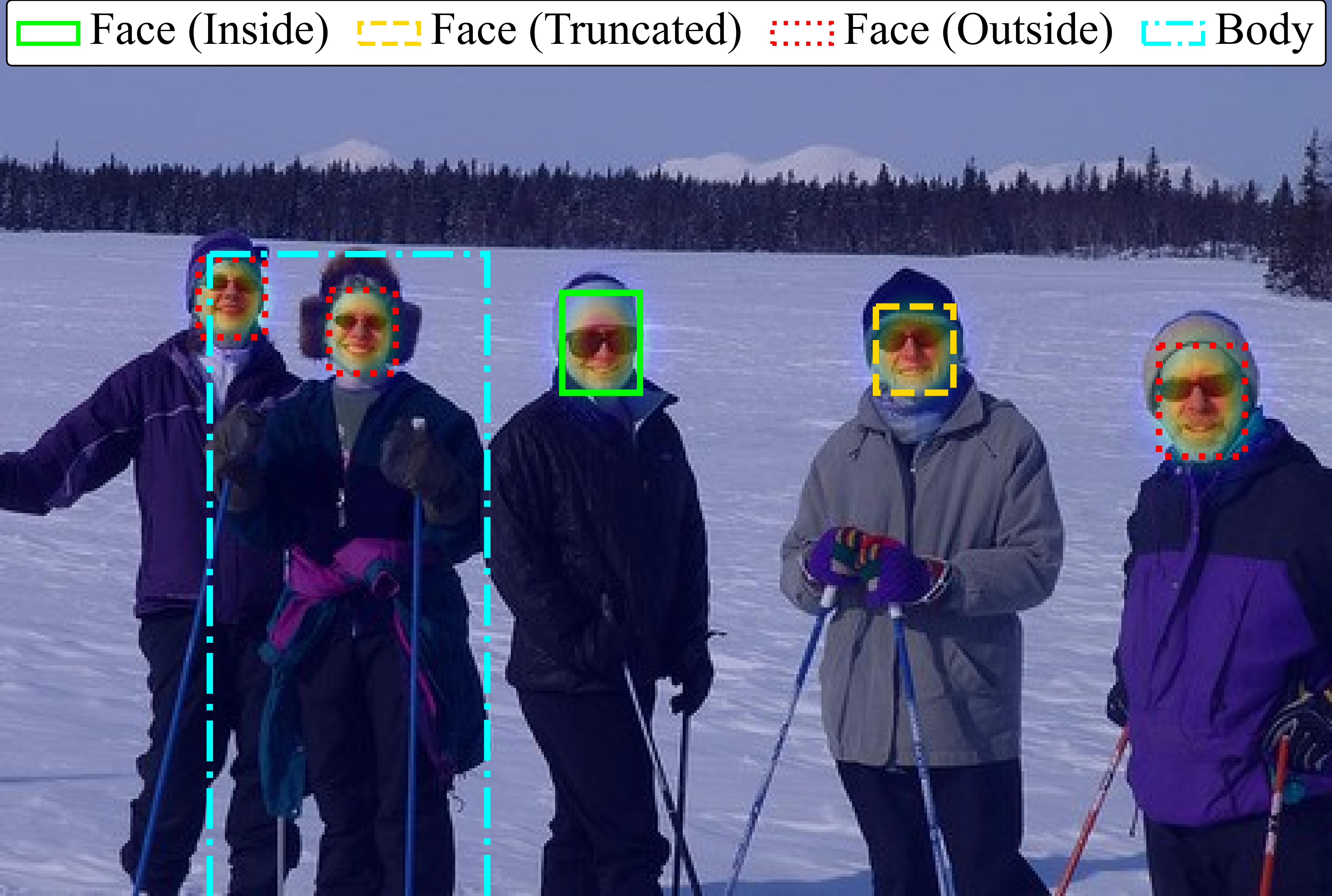}
        \caption{Ground-truth annotations}
        \label{fig:task_c}
    \end{subfigure}%
    \vspace{-6pt}
    \caption{
        Extreme amodal face detection.
        This task predicts, given an input image, the likelihood of faces at all locations within an expanded field-of-view frame.
        Specifically, a face presence heatmap and bounding boxes are estimated both inside and outside the image.
        In the example pictured, there is direct visual evidence of three faces (one in-frame, one partially in-frame, and one with a partially-observed correlate---the person's body), and two more faces without direct evidence but with a non-zero conditional probability.
    }
    \label{fig:task}
}

\makeatletter
\apptocmd\@maketitle{{\par\vspace{0pt}\splashfigure{}\par\vspace{14pt}}}{}{}
\makeatother

\begin{document}

\maketitle




\begin{abstract}

Extreme amodal detection is the task of inferring the 2D location of objects that are not fully visible in the input image but are visible within an expanded field-of-view.
This differs from amodal detection, where the object is partially visible within the input image, but is occluded.
In this paper, we consider the sub-problem of face detection, since this class provides motivating applications involving safety and privacy, but do not tailor our method specifically to this class.
Existing approaches rely on image sequences so that missing detections may be interpolated from surrounding frames or make use of generative models to sample possible completions.
In contrast, we consider the single-image task and propose a more efficient, sample-free approach that makes use of the contextual cues from the image to infer the presence of unseen faces.
We design a heatmap-based extreme amodal object detector that addresses the problem of efficiently predicting a lot (the out-of-frame region) from a little (the image) with a selective coarse-to-fine decoder.
Our method establishes strong results for this new task, even outperforming less efficient generative approaches. Code, data, and models are available at \url{https://charliesong1999.github.io/exaft_web/}.


\end{abstract}
    
\section{Introduction}
\label{sec:intro}

Object detection has been a central problem in computer vision for decades~\cite{od_survey,open_od_survey}, with significant advances in closed-set detection of predefined categories~\cite{od_survey} and open-set detection that generalizes beyond fixed taxonomies~\cite{open_od_survey}.
However, existing detectors are fundamentally constrained to objects visible within the input frame. 
This restricts their applicability in scenarios
that require extrapolating beyond what is directly observable.

We take a step toward this broader goal by introducing the task of \emph{extreme amodal detection}, where the objective is to detect and localize objects that may lie partially or entirely outside the visible field-of-view. 
While our design is applicable to the general task, in this paper we focus on the sub-problem of extreme amodal \textit{face} detection, which is especially well-motivated due to its relevance to safety-critical (\eg, anticipating pedestrians), accessibility-related (\eg, assisting those with visual impairments \cite{blind_people}), and privacy-sensitive applications. 
As shown in \cref{fig:task}, we categorize extreme amodal faces into
(1)~\emph{truncated faces}, which are partially within the field-of-view; and
(2)~\emph{outside faces}, where the face is completely outside the field-of-view.
The latter is subdivided into two cases:
(2a)~\emph{with evidence}, where direct visual evidence, such as a visible body, is observed; and
(2b)~\emph{without evidence}, where the model must rely on indirect contextual cues. 


The impact on privacy is especially relevant, and worth elaborating as it explains our focus on human faces in particular.
In brief, extreme amodal face detection can improve privacy by enabling computer vision systems to actively avoid capturing sensitive information, \ie, human faces.
Cameras in public spaces pose inherent privacy risks, and cameras that move in public spaces (\eg, on self-driving cars, drones, or other semi-autonomous robotic systems) exacerbate those risks. 
Existing solutions often aim to secure data during post-processing, for example, by detecting and blurring faces.
This is not a robust strategy, however, as raw data is susceptible to theft \cite{mike6,mike12,mike13}, corporate misuse \cite{mike7,mike8,mike9}, or legally enforced retrieval \cite{mike10,mike11}.
More fundamentally, this strategy overlooks data collection as a site of intervention, and that the best privacy-preserving strategy is often to not collect sensitive data at all.
Extreme amodal face detection can serve this end.
If deployed successfully, it can limit the need for actual surveillance, preserving privacy without sacrificing utility.
%

Prior work provides only limited tools for this task. 
Tracking-based methods~\cite{tao-amodal} leverage temporal continuity in video to recover partially unseen objects, but do not address the case of a single static frame. 
Another line of work relies on generative pipelines that outpaint the extended frame using diffusion-based models~\cite{pix2gestalt, bhattacharjee2025believing}, followed by conventional detectors. 
While straightforward, these approaches have several drawbacks: 
(a)~they depend heavily on additional prompts (\eg, text or masks) whose quality can significantly affect the results; 
(b)~diffusion models are computationally expensive and slow at inference time, making them ill-suited for time-critical detection scenarios; and
(c)~these pipelines are not end-to-end trainable, limiting their ability to adapt to new detection tasks. 
In contrast, humans can readily infer the existence and location of unseen objects based on prior knowledge, contextual cues, or reasoning from visible body parts. 

The extreme amodal setting introduces three unique challenges. 
First, the extended region can, in principle, be arbitrarily larger than the input image. 
In our setup, we restrict this extension to $8\times$ the input size, which nonetheless requires the long-distance extrapolation of information.
Second, naively querying the entire extended region is computationally prohibitive, requiring up to $8\times$ more tokens and wasting resources on regions that often contain no objects.
Third, the underlying true conditional distribution cannot be accessed; we only have a single realization for any input image.
This poses a challenge for evaluation, where we can only measure success indirectly using the ground-truth realization, as discussed in \cref{sec:experiment}.

To address the first two issues, we propose a \emph{coarse-to-fine selective decoder} that makes good use of limited information while remaining compute-efficient. 
Our decoder first queries the extended area at low resolution, dividing it into candidate regions. 
It then selectively refines only a subset of promising candidates to match the resolution of the input image. 
This design reduces the number of tokens by lowering the resolution at the initial stage and by refining only the most relevant candidates. 
As a result, our approach achieves both efficiency and strong detection performance.
Our contributions are threefold. We
\begin{enumerate}
    \item introduce extreme amodal face detection, the task of detecting and localizing faces partially or entirely outside the visible field-of-view;
    \item construct a benchmark dataset derived from COCO~\cite{coco-dataset} images, enabling systematic evaluation for faces inside the image, outside the image, and truncated by the image frame; and
    \item design an efficient and effective extreme amodal detector with a novel coarse-to-fine selective decoder.
\end{enumerate}

\section{Related Work}
\label{sec:related_work}


Existing works related to our task can be broadly grouped into two categories: \emph{tracking-based approaches}, which leverage temporal information across multiple frames, and \emph{generative-based approaches}, which rely on large generative models conditioned on additional prompts such as masks or text.

\paragraph{Tracking-based methods.}
OccludTrack~\cite{van2023tracking} introduced the problem of tracking objects even when fully invisible, either due to occlusion or containment. 
Their dataset was collected via simulation and manual labeling. Cotracker~\cite{karaev2024cotracker} extended point tracking by jointly tracking all points, demonstrating strong robustness in fully occluded and out-of-frame scenarios. 
TAO-amodal~\cite{tao-amodal} expanded bounding boxes of pre-trained trackers beyond the visible frame by exploiting temporal consistency. 
ObjectRemember~\cite{objectRemember} lifted object points into 3D coordinates, storing them in memory to persist objects even when they leave the frame. 
While these methods can estimate truncated or outside objects, they inherently require temporal cues across multiple frames. 
In contrast, our work investigates how to detect such objects from a \emph{single} static frame.

\paragraph{Generative-based methods.}
A second line of work uses generative models to complete or outpaint missing regions. 
In amodal completion, Pix2Gestalt~\cite{pix2gestalt} employed SAM~\cite{kirillov2023segment} to obtain masks and fine-tuned a diffusion model for part-whole completion. 
PD-MC~\cite{xu2024amodal} used grounded-SAM~\cite{ren2024grounded} with text prompts to automatically generate masks, then progressively completed objects. 
OpenACC~\cite{ao2025open} further incorporated both masks and background context to reason about text prompts for flexible completion. 
These methods, however, primarily address the occlusion problem but do not necessarily generalize well to cases where objects of interest are truncated or completely invisible. However, our method can infer completely invisible objects. 

For outpainting, PQ-Diff~\cite{zhang2024continuousmultiple} trained a diffusion model with positional queries for arbitrary-size extrapolation, though performance degrades in complex scenes. 
VIP~\cite{VIP} employed large multimodal models to provide semantic supervision during outpainting. 
Unseen~\cite{bhattacharjee2025believing} generated the unseen regions with additional text prompts before applying a detector. 
Despite their creativity, generative-based pipelines share key drawbacks: they rely on external prompts (mask or text), require large diffusion models that are computationally expensive and slow at inference, and are not end-to-end trainable. 
These limitations make them unsuitable for fast and efficient detection scenarios, such as detecting out-of-frame pedestrians in autonomous driving.

\section{Extreme Amodal Detection}
\label{sec:task_definition}

Given an image $\bm{x} \in \reals^{H \times W \times 3}$, extreme amodal detection predicts the location of objects within a centrally-expanded region of size $KH \times KW$, where $K$ denotes the expansion factor.
To predict objects within this larger region, we consider two output types, commonly associated with the tasks of detection and localization.
For the detection task, a set of $N$ objects $o_i = (c_i, b_i)$ are predicted, where $c_i$ denotes the object class and $b_i = (x_i, y_i, w_i, h_i)$ denotes the bounding box represented by center coordinate, width and height.
%
For the localization task, a heatmap $\bm{h} \in [0,1]^{KH \times KW \times C}$ is predicted, where $C$ denotes the number of classes, indicating the probability that an object of each class is located at that pixel.
As motivated in the introduction, in this paper we consider a single class: human faces.


As shown in \cref{fig:task}, the difficulty of detecting extreme amodal faces varies, depending on whether there is direct visual evidence within the image of a face wholly or partially outside the image.
We classify faces as
\begin{enumerate}
    \item \textbf{Inside:} faces that are entirely within the image;
    \item \textbf{Truncated:} faces that are partially within the image; and
    \item \textbf{Outside:} faces that are entirely outside the image,
    \begin{enumerate}
        \item with direct visual evidence, such as a visible body in the image; and
        \item without direct visual evidence, where indirect cues like eye gaze and semantic co-occurrences may need to be considered.
    \end{enumerate}
\end{enumerate}

\section{The \dataset Dataset}
\label{sec:dataset}

\begin{table}[!t]
\centering
\setlength{\tabcolsep}{1pt}
\begin{tabularx}{\linewidth}{@{}lccCC@{}}
    \toprule
    $\# \times 10^3$ (\%) & Inside & Truncated & Outside + & Outside -\\
    \midrule
    Boxes (train) & 116 (24\%) & 74 (15\%) & 66 (13\%) & 235 (48\%) \\
    Boxes (test) & 5.0 (24\%) & 3.0 (14\%) & 2.0 (12\%) & 11 (50\%) \\
    \midrule
    Images (train) & 30 (17\%) & 37 (20\%) & 32 (17\%) & 83 (46\%) \\
    Images (test) & 1.0 (16\%) & 1.5 (20\%) & 1.0 (17\%) & 3.5 (47\%) \\
    \bottomrule
\end{tabularx}
\caption{
\dataset dataset statistics.
Sample counts ($\times 10^3$) and percentages are shown for bounding boxes and images.
The data is divided into subsets of inside faces, truncated faces, outside faces with direct evidence (+), and outside faces without direct evidence (-).
The category of an image is determined by its hardest face.
}
\label{tab:method-statistic-data}
\end{table}

In this section, we introduce the Extreme Amodal Face (\dataset) dataset, derived from the MS COCO~\citep{coco-dataset} object detection dataset.
First, RetinaFace~\citep{retinaface} was used to pseudolabel the many unlabeled faces in the COCO dataset, excluding those detections with a confidence below 0.9, resulting in 2.4$\times$ more face labels.
Next, the images were randomly cropped and the bounding boxes from the cropped and uncropped regions were retained.
For an image with height $H$ and width $W$, the process is as follows.
\begin{enumerate}
    \item Randomly sample crop height from $[0.3H, 0.6H]$ and aspect ratio from $[0.5, 2]$, yielding the crop size $H' \times W'$.
    
    \item Randomly sample center $x$ coordinate from $[0.5W' , W - 0.5W']$ and $y$ coordinate from $[0.5H', H - 0.5H']$.

    \item Crop image using crop size and center.
    
    \item Discard bounding boxes that are not fully contained within an expanded area $K^2\times$ the size of the crop.
    
    \item Update the bounding box center coordinates $(x_b, y_b)$ to the expanded image coordinate frame: $(x_b - x + 0.5KW', y_b - y + 0.5KH')$.
\end{enumerate}
This is repeated 4 times per image to generate diverse data.
The dataset statistics are given in \cref{tab:method-statistic-data}.
%



\section{Extreme Amodal Face Detector}
\label{sec:method}




\begin{figure*}
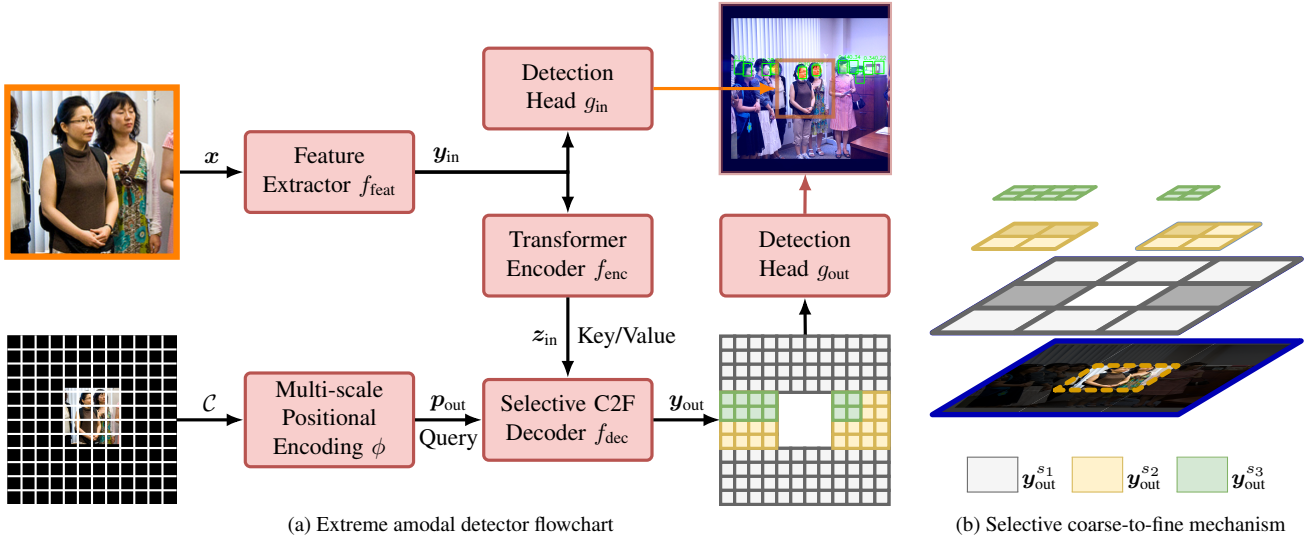

  \centering
  \begin{subfigure}{0.68\linewidth}
    \includegraphics[width=\linewidth]{img/overall.tikz}
    \caption{Extreme amodal detector flowchart}
    \label{fig:overall}
  \end{subfigure}\hfill
  \begin{subfigure}{0.3\linewidth}
    \includegraphics[width=\linewidth]{img/C2F.tikz}
    \caption{Selective coarse-to-fine mechanism}
    \label{fig:sc2f}
  \end{subfigure}%
  \vspace{-6pt}
  \caption{
  Overview of our extreme amodal detector.
  (a)~Flowchart of our approach.
  Given an input image, a feature map is extracted, from which a dedicated in-image detection head infers object boxes and a face probability heatmap.
  Separately, a transformer encoder--decoder shares information from the image to the extended area around the image.
  We propose an efficient selective coarse-to-fine decoder that starts with low resolution out-of-image positional encodings as the input tokens, then refines a selected subset of these tokens at higher resolutions.
  A second detection head uses these tokens to infer the out-of-image object boxes and heatmap.
  (b)~Illustration of our selective coarse-to-fine mechanism.
  We first query the low-resolution regions, then use a scoring network to rank these regions and select the top-$\mu$\% to be refined at a higher resolution, until at the same resolution as the input image feature map.
  }
  \label{fig:framework_and_C2F}
\end{figure*}

In this section, we outline our extreme amodal face detector, as shown in \cref{fig:framework_and_C2F}.
Our method involves feature extraction, a transformer encoder--decoder for sharing information between in-image tokens and out-of-image tokens, and two detection heads, one for in-image faces and one for out-of-image faces.
First, a convolutional feature extractor $f_\text{feat}$ computes a feature map $\bm{y}_\text{in}$ given the image.
Then, a transformer encoder $f_\text{enc}$ processes these features into a form useful for predicting out-of-image faces, given rotary positional encodings $\bm{p}_\text{in} = \phi(\cC_\text{in})$ of the in-image coordinates $\cC_\text{in}$ \cite{heo2024rotary}.
Next, our selective course-to-fine transformer decoder $f_\text{dec}$ cross-attends to the in-image features, given the positional encodings $\bm{p} = \phi(\cC)$ of the expanded image coordinates $\cC$.
Finally, two detection heads $g$ predict in- and out-of-image objects $o$ and heatmaps $\bm{h}$.
In summary, we have
\begin{align}
    \bm{y}_\text{in} &= f_\text{feat}(\bm{x})\\
    \bm{z}_\text{in} &= f_\text{enc}(\bm{y}_\text{in}, \bm{p}_\text{in})\\
    \bm{y}_\text{out} &= f_\text{dec}(\bm{z}_\text{in}, \bm{p}) \label{eq:dec}\\
    (o_\text{in}, \bm{h}_\text{in}) &= g_\text{in}(\bm{y}_\text{in})\\
    (o_\text{out}, \bm{h}_\text{out}) &= g_\text{out}(\bm{y}_\text{out}).
\end{align}
The main novelty of the approach arises from the transformer decoder, which will now be outlined in detail.

\paragraph{Selective course-to-fine (C2F) decoder.}
Sharing information between the image and the extended region beyond the image is challenging for two reasons:
(a)~high computational cost: if using the same resolution, the extended region has $(K^2-1)\times$ more tokens than the input image; and
(b)~object sparsity: only a small proportion of image patches contain objects (in our dataset, fewer than 1\% of the $16\times 16$ pixel patches contain faces).
However, it is not possible to know which patches contain objects in advance.
To address this, we propose a selective coarse-to-fine mechanism: first query the extended region at low resolution, then use a scoring network to select promising regions for refinement.

The approach is as follows.
As indicated in \cref{eq:dec}, the transformer decoder receives the in-image features $\bm{z}_\text{in}$, which are projected into keys and values, and the positional encodings $\bm{p}$.
For the first decoder layer, low-resolution, coarse positional encodings $\bm{p}_\text{out}^{s_1}$ from the extended region around the image are projected into queries.
The positional encodings are given by
\begin{align}
    \bm{p}_\text{out}^{s_i} = \{ \text{avgpool}(\bm{p}, s_i) (u, v) \mid (u,v) \in \cC_\text{out} \},
    \label{eq:avg_pool}
\end{align}
where $\text{avgpool}(\cdot, s_i)$ denotes average pooling with an $s_i \times s_i$ window, $\cC_\text{out}$ denotes the out-of-image coordinates, and $s_i \in \mathcal{S}$ are a sequence of coarse-to-fine scales.
The decoder layer uses ROPE positional encodings \cite{heo2024rotary} to facilitate cross-attention between within-image and out-of-image tokens at the requisite scale.
After the first 2-layer decoder block $f_\text{decblk}$, a scoring network $f_\text{score}$ predicts which tokens to refine at a higher resolution, retains only the top-$\mu^{s_i}\%$ tokens, and duplicates these to match the number required by the next resolution level.

In summary, initialization sets
$\bm{x}_\text{out}^{s_1} \gets \bm{p}_\text{out}^{s_1}$
and then the per-block computations proceed as 
\begin{align}
    \bm{y}_\text{out}^{s_i} &= f_\text{decblk}(\bm{x}_\text{out}^{s_i}, \bm{p}_\text{out}^{s_i}, \bm{z}_\text{in}, \bm{p}_\text{in})\\
    \bm{x}_\text{out}^{s_{i+1}} &= f_\text{score}(\bm{y}_\text{out}^{s_i}, \mu^{s_i}).
\end{align}
The output features at each scale are aggregated by summing upsampled (if necessary) feature maps,
\begin{align}
    \bm{y}_\text{out} &= \sum_{i=1}^{|\cS|} \uparrow(\bm{y}_\text{out}^{s_i}).
\end{align}

\section{Experiments}
\label{sec:experiment}

\begin{table*}[!t]
\centering
\setlength{\tabcolsep}{2pt}
\begin{tabularx}{\textwidth}{@{}l C>{\columncolor[gray]{0.9}}cCCC c>{\columncolor[gray]{0.9}}cccc c>{\columncolor[gray]{0.9}}c>{\columncolor[gray]{0.9}}c>{\columncolor[gray]{0.9}}c@{}}
    \toprule
    Method &
    AP$\uparrow$ & AP\textsubscript{t}$\uparrow$ & AP\textsubscript{o}$\uparrow$ & AP\textsubscript{o+}$\uparrow$ & AP\textsubscript{o-}$\uparrow$ &
    MAE$\downarrow$ & MAE\textsubscript{t}$\downarrow$ & MAE\textsubscript{o}$\downarrow$ & MAE\textsubscript{o+}$\downarrow$ & MAE\textsubscript{o-}$\downarrow$ &
    mIoU\textsubscript{o}$\uparrow$ & 
    AR\textsubscript{o}$\uparrow$ & 
    SE\textsubscript{o}$\downarrow$ & 
    CE\textsubscript{o}$\downarrow$\\
    \midrule
    Uniform & -- & -- & -- & -- & -- & -- & -- & -- & -- & -- &8.80&51.71& 100 & 100\\
    Oracle-GT & 100 &100&100&100&100&0.00&0.00&0.00&0.00&0.00& 100 &100& 58.68 & 58.68\\
    Oracle-YOLOH & 44.79 & 61.70 & 36.34 & 49.83 & 22.85 & 7.55 & 2.07 & 10.65 & 2.54 & 13.60 & 28.63 & 44.56 & 91.96 & 78.74 \\
    \midrule
    YOLOH \cite{yoloh} & 10.20 & 30.60 & 0.01 & 0.01 & 10$^{-3}$ & \underline{17.37} & 2.78 & 26.11 & 6.87 & \underline{33.11} & 17.23 & 19.01 & 96.90 & 94.01 \\
    Pix2Gestalt \cite{pix2gestalt} & \underline{11.30} & \underline{33.43} & 0.24 & 0.48 & 10$^{-3}$ & 17.38 & 2.83 & \underline{26.10} & 6.63 & 33.18 & 17.75 & 20.25 & 96.54 & 93.31 \\
    Outpaint \cite{podell2023sdxl} & 4.93 & 11.54 & \textbf{1.62} & \textbf{2.47} & \textbf{0.76} & \textbf{14.69} & 2.07 & \textbf{21.94} & \textbf{3.48} & \textbf{28.67} & \textbf{20.53} & \underline{25.03} & \underline{96.41} & \underline{90.18} \\
    Ours & \textbf{23.07} & \textbf{66.69} & \underline{1.26} & \underline{2.17} & \underline{0.34} & 17.83 & \textbf{2.01} & 27.43 & \underline{4.53} & 35.77 & \underline{18.70} & \textbf{27.17}  & \textbf{93.99} & \textbf{88.16} \\
    \bottomrule
\end{tabularx}
\caption{
Extreme amodal detection performance on the test set of our MS COCO-based dataset.
We report
the average precision (AP),
the mean absolute error (MAE) of the nearest bounding box center,
the mean intersection-over-union (mIoU),
the average recall (AR), 
the self-entropy (SE), and
the cross-entropy (CE).
The data subsets truncated (t), outside (o), outside with evidence (o+), and outside without evidence (o-) are indicated by subscripts.
The metrics that are most meaningful for assessing performance on the different data subsets are shaded.
Detection metrics like AP are appropriate for evaluation of the truncated faces, since the realization of the conditional distribution (our ``ground-truth'') is very close to the true distribution near the image.
However, further from the image, this realization no longer captures all modes of the true distribution, and so AR, CE and SE are more meaningful measures of performance in this regime.
}
\label{tab:main_result}
\end{table*}

\begin{table}
\centering
\setlength{\tabcolsep}{2pt}
\footnotesize
\begin{tabularx}{\linewidth}{@{}lCCCCcc@{}}
    \toprule
     & 
     \#Params&
     Memory&
     FLOPs &
     Latency&
     Throughput&
     VRAM
     \\
     Method & 
     $\times 10^6$&
     (MB)&
     $\times 10^9$&
     (ms)&
     (s$^{-1}$) &
     (MB)
     \\
    \midrule
    YOLOH & \textbf{42.8} & \textbf{164} & \textbf{20.4} & \textbf{9.0} & \textbf{111.9} & \textbf{428} \\
    Pix2Gestalt & 3.5k & 7k & 452k & 7.2k & 0.3 & 31k \\
    Outpaint & 7.3k & 14k & 467k & 7.4k & 0.1  & 31k \\
    Ours & \underline{67.8} & \underline{259} & \underline{47.6} & \underline{161.6} & \underline{6.2} & \underline{728} \\
    \bottomrule
\end{tabularx}
\caption{
Inference efficiency on a single L40S GPU. 
We report the number of parameters, the memory size of the parameters, the computational cost, the latency at the 95th percentile, throughput in iterations per second, and peak VRAM usage.
Generative pipelines (Pix2Gestalt and Outpaint) require orders of magnitude more parameters and FLOPs, resulting in prohibitive latency and memory consumption. 
}
\label{tab:efficiency}
\end{table}

\begin{figure*}[!t]
    \centering
    \includegraphics[width=\textwidth]{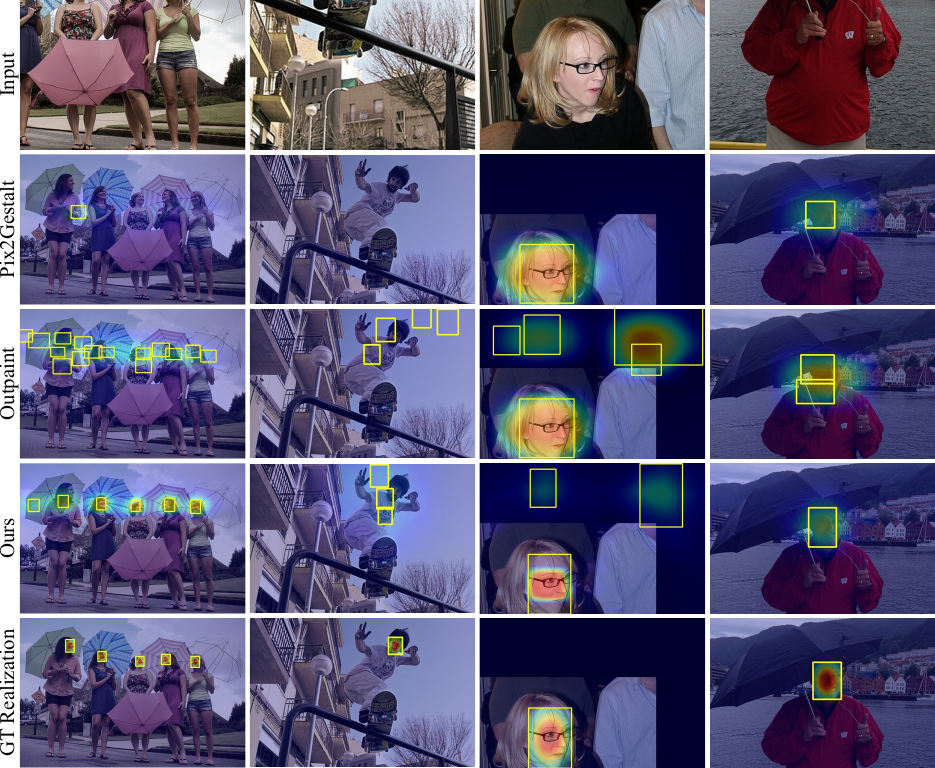}
    \caption{
    Qualitative results. 
    The final row shows samples from the ground-truth conditional distributions. 
    Our model effectively leverages contextual cues—such as nearby people (example 1), objects like a skateboard (example 2), or partial body evidence (example 4)—to infer completely unseen faces. 
    In example 1, the model correctly extends predictions to the left, where a partial person is visible, but not to the right, demonstrating awareness of scene context and typical human height. 
    Example 3 highlights the model's generalization to real-world scenarios. 
    Unlike other examples where inputs are synthetically cropped from complete images, this example is naturally truncated (i.e., the faces were never captured in the original photo). 
    Our model successfully generates plausible faces despite the lack of ground truth, demonstrating its practical utility for real-world photo expansion.
    Compared to our model, Pix2Gestalt struggles without large visible body parts, while the outpainting pipeline can infer outside faces but yields noisier and less consistent results.
    }
    \label{fig:qualitative_result}
\end{figure*}

In this section, we evaluate our approach on our EXAFace dataset and compare it with an object detector baseline and two generation-based methods.
Our method outperforms all compared approaches while being significantly more efficient than those that require image generation.
We also analyze our design choices and report failure cases.


\subsection{Experiment setup}

\paragraph{Detection metrics.}
Average precision (AP) and mean absolute error (MAE) are reported to evaluate the accuracy of the predicted bounding boxes.
AP is given at a 25\% intersection-over-union (IoU) threshold, a looser threshold than is used for the standard detection task since extreme amodal detection is considerably more challenging.
MAE measures how far the predicted object centers are from the ground-truth centers, where predictions and ground-truth centers are paired using the Hungarian algorithm.
We report the MAE normalized by the diagonal of the input image so that it is independent of the image resolution.
Since we necessarily evaluate with respect to a realization of the ground-truth conditional distribution, these metrics are only reliable measures close to the input image, where the realization approximates the conditional distribution.
Therefore, they are suitable only for evaluating truncated faces.

\paragraph{Localization metrics.}
Heatmap IoU, average recall (AR), cross-entropy (CE), and self-entropy (SE) are reported to evaluate the accuracy of the predicted heatmaps outside of the image. 
Since we evaluate with respect to a realization of the true distribution, AR, CE and SE are the most relevant metrics for assessing performance.
That is, a prediction that has modes in addition to those of the observed sample of the ground-truth distribution should still be considered good, and this can be assessed using AR and CE.
Equally, it is important to check that the prediction is not uniform by consulting the self-entropy.

\paragraph{Compared methods.}
We compare our method with three baselines/oracles and three state-of-the-art approaches.
The baselines include
a uniform heatmap prediction (Uniform), where the presence of a face is set to be equally likely at all locations in the expanded region;
an oracle that yields the ground-truth realization (Oracle-GT); and
an oracle that applies the YOLOH object detector \cite{yoloh} to the real extended image (Oracle-YOLOH).
The compared methods include
YOLOH \cite{yoloh}, given a black-padded input image the size of the required output;
Pix2Gestalt \cite{pix2gestalt}, a method that amodally completes partially occluded bodies, given the ground-truth in-image masks, resulting in an extended image that is passed to the YOLOH detector; and
Outpainting, similar to \citet{bhattacharjee2025believing}, where a diffusion model generates many samples of outpainted images, with text prompts generated by a vision--language model (VLM), which are passed to the YOLOH detector whose predictions are aggregated.
The diffusion model and VLM used for this model are almost certain to have seen the extended images in our test set.
Note that all methods use the same YOLOH detector that we trained on our dataset to predict bounding boxes and heatmaps of faces and bodies.


\paragraph{Implementation details.} 

Our extreme amodal detector extends the pre-trained YOLOH~\cite{yoloh} detector's feature extractor and detection head with a two-layer transformer encoder and a two-layer selective C2F transformer decoder. 
Transposed convolutions are used for upsampling, and the scoring network shares the same architecture as the YOLOH detection head.
The expansion ratio is $K=3$ and the multi-scale refinement set is $S=(2,1)$.
Input images are resized to $320\times320$ and normalized, without additional augmentation.  
The model is optimized with AdamW, with the momentum parameter set to $0.9$, weight decay set to $10^{-2}$, and learning rates set to $0.024$ for the transformer and detection head, and $0.004$ for the YOLOH backbone. 
We use a warm-up scheduler~\cite{vaswani2017attention}, where the learning rate is scaled with the embedding dimension and number of warm-up steps (20\% of the total). 
A decay factor of $0.1$ is applied after 100 steps. 
The model is trained for 14 epochs on four A100 GPUs with a batch size of 64. 
For the ablation study, we train for 8 epochs on 25\% of the EXAFace dataset with a batch size of 32 on two 2080Ti GPUs.

The baseline YOLOH detector with a dilated ResNet-50 backbone and a CNN-based decoder is trained for 14 epochs on pseudo-labeled COCO~\cite{coco-dataset} faces and bodies to predict bounding boxes and heatmaps. 
The input resolution is $320\times320$, random horizontal flip and random shift augmentations are applied, the learning rate is $0.03$, the warm-up iterations are 1200, step decays of $0.003$ and $0.0003$ at the 8th and 11th epochs are applied, and the batch size is 32 on two 2080Ti GPUs.
For the generative baselines, we use the official Pix2Gestalt~\cite{pix2gestalt} checkpoint, following the gradual completion strategy of~\cite{xu2024amodal}. 
Masks touching image boundaries are iteratively extended by 10\% until completion, and multiple bodies are completed sequentially and merged. 
For the outpainting pipeline, BLIP2~\cite{li2023blip} generates text captions as prompts, which are fed into SDXL~\cite{podell2023sdxl} for image extrapolation.

\subsection{Results}

\begin{table}
\footnotesize
\begin{tabularx}{\linewidth}{@{}lCCC@{}}
    \toprule
     Method                     &
     AP\textsubscript{t}$\uparrow$       &
     MAE\textsubscript{t}$\downarrow$    &
     AR\textsubscript{o}$\uparrow$                  \\
    \midrule
    Ours & 62.73 & 2.19 & 26.64\\
    w/o average pooling & 61.13 & 2.28 & 25.67\\
    w/o multi-scale & 61.28 & 2.25 & 25.24 \\
    \bottomrule
\end{tabularx}
\caption{Ablation study. 
Here, ``w/o average pooling'' replaces average pooling with center sampling for downsampling the positional encodings, and
``w/o multi-scale'' restricts the decoder to a single scale. 
Both components improve performance across all three metrics: 
average pooling contributes more to bounding box localization (AP$_\text{t}$, MAE$_\text{t}$), while the multi-scale selective C2F mechanism yields greater gains in heatmap quality (AR$_\text{o}$).
}
\label{tab:ablation}
\end{table}

\begin{figure}[!t]
    \centering
    \includegraphics[width=\linewidth]{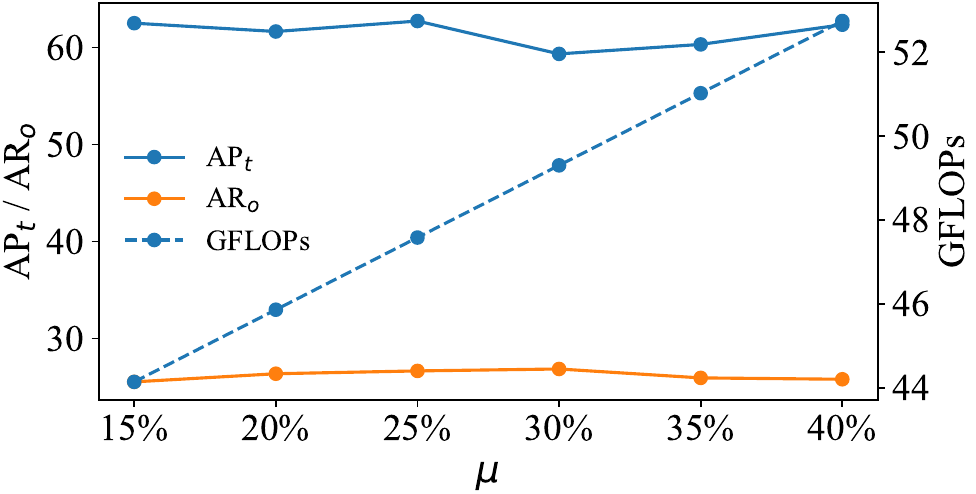}
    \caption{Sensitivity analysis of the percentage of retained tokens $\mu$ at scale $\cS = (2)$. 
    The metrics are relatively insensitive to $\mu$, so we select $\mu=25\%$, which is computationally efficient without sacrificing performance.
    The original data is shown in the appendix (\cref{tab:ana_topu2}).
    }
    \label{fig:top_mu2}
\end{figure}

\begin{figure}[!t]
    \centering
    \includegraphics[width=\linewidth]{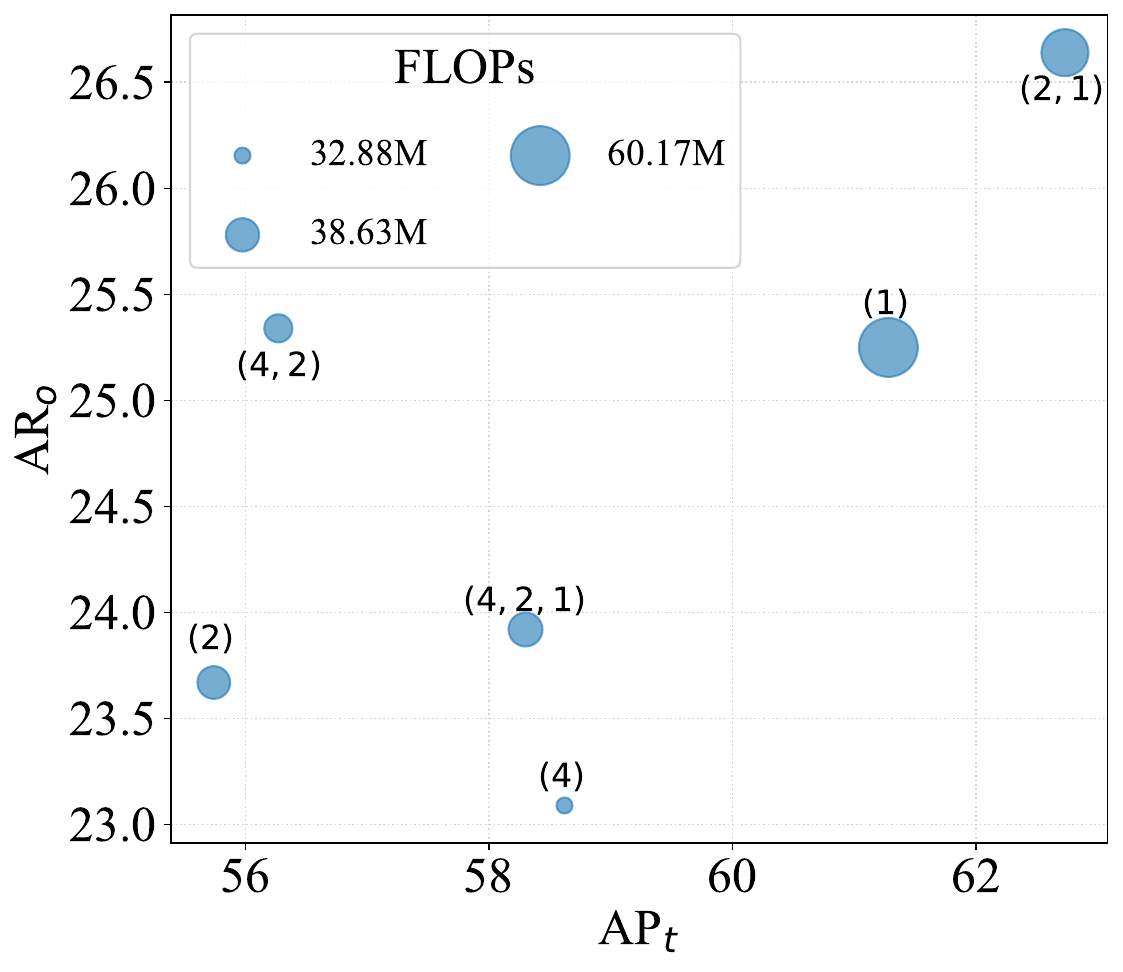}
    \caption{Analysis of multi-scale settings.
    We evaluate three scales $s=1,2,4$ and their combinations $\cS = (4,2)$, $(2,1)$, $(4,2,1)$. 
    The results show that $\cS=(2,1)$ yields the highest AP$\textsubscript{t}$ and AR$\textsubscript{o}$, and is therefore adopted as our default setting.
    Original data is shown in the appendix (\cref{tab:ana_multiscale}).
    }
    \label{fig:ana_multi-scale}
\end{figure}

\begin{figure*}[!t]
    \centering
    \includegraphics[width=\textwidth]{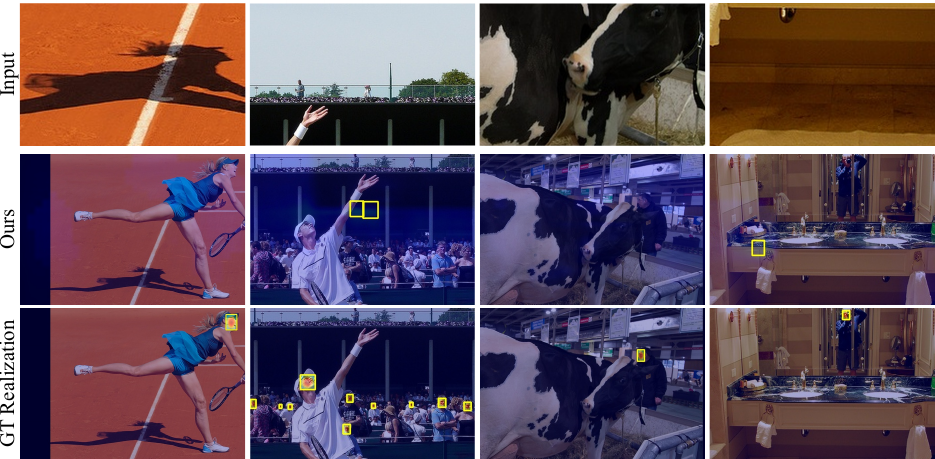}
    \caption{
    Failure cases. 
    Our model struggles to predict outside faces when contextual cues are weak. In the first and second examples, strong appearance evidence is present but location cues are limited. 
    In the third and fourth examples, no appearance evidence is available, making the presence and location of an outside face ambiguous—even for human observers.
    }
    \label{fig:failure_cases}
\end{figure*}

Quantitative and qualitative results are given in \cref{tab:main_result} and \cref{fig:qualitative_result}, respectively.
Our model consistently outperforms all comparison methods, while also having significantly better inference efficiency than generative methods (\cref{tab:efficiency}). 
It is important to note that since we evaluate the performance on a \textit{realization} of the ground-truth conditional distribution, AP and MAE are not suitable for measuring the detection performance outside the image, though they are appropriate for truncated faces where the true realization and true distribution overlap.
For faces outside the image frame, heatmap metrics like average recall and cross-entropy are more suitable, since they do not punish the prediction of additional modes beyond those contained in the realization, unlike the mIoU metric.
This is desirable because the true conditional distribution is likely to have more modes than a realization: there are multiple possible plausible configurations.
However, these metrics should be considered in parallel with self-entropy to verify that the model is not predicting a near-uniform distribution, which is also implausible.
In \cref{tab:main_result}, we shade the columns that are most meaningful for assessing performance on this task.
Our approach exhibits a strong ability to predict face locations, whether or not there is direct visual evidence.

The outpainting pipeline also achieves strong alignment with the realized ground-truth distribution of outside faces, outperforming our approach on AP\textsubscript{o} and MAE\textsubscript{o}, albeit with $10000\times$ the FLOPS.
While these metrics are not suitable for measuring performance with respect to the true distribution, they should also be interpreted with some caution regardless: there is very likely information leakage, since BLIP2~\cite{li2023blip} is trained on COCO and SDXL~\cite{podell2023sdxl} is likely to have been trained on COCO.
Therefore, the model is almost certain to have seen the extended images in our test set.
A visual example of outpainting is shown in the appendix (\cref{fig:ex_outpaint}).
%
In contrast, Pix2Gestalt~\cite{pix2gestalt} often fails to amodally complete the truncated part of the face.
This is expected, since the model is trained for in-frame occluder removal, not for occlusions caused by the camera's field-of-view.
A visual example of a completion by Pix2Gestalt is shown in the appendix (\cref{fig:ex_pix2gestalt}).
Finally, it is interesting that our approach outperforms the YOLOH oracle that receives the extended image for truncated faces.
This is attributable to the input resolution: both methods process a $320 \times 320$ image, but the resolution of the cropped region is effectively higher for our approach.
%
%

\subsection{Ablation study and analysis}
\label{sec:ablation}

In \cref{tab:ablation}, we ablate two design choice:
the positional encoding downsampling strategy of average pooling is replaced with center sampling, and
the multi-scale decoding strategy is replaced by a single scale.
The results indicate that replacing either of these design choices with simpler approaches leads to significantly poorer performance.


\cref{fig:ana_multi-scale} presents the analysis of different multi-scale strategies.
Among the explored settings, the $(2,1)$ configuration achieves the best overall performance, and is therefore adopted as our default.
\cref{fig:top_mu2} shows the effect of varying $\mu$, where it is clear that the metrics are relatively insensitive to this hyperparameter choice.
This confirms that our selection mechanism is computationally advantageous without sacrificing accuracy.


\subsection{Limitations and discussion}

Several failure cases of our method are shown in \cref{fig:failure_cases}.
This highlights one limitation of our approach, that it struggles when the contextual cues are weak, such as a person's shadow but no body.
This may stem from insufficient training data to capture such rare examples, or from the inherent ambiguity in these scenarios. 
Another limitation is that our approach predicts the conditional distribution of a face outside the image, but cannot be used to sample multiple co-occurring faces.
In contrast, the outpainting method samples co-occurring faces and so retains these useful correlations.
This may limit the use of our approach in some downstream applications, where we may wish to know about the plausible configurations of multiple objects.
A final limitation is that we have only considered the class of human faces.
However, our approach is not tailored specifically to faces, and should easily extend to other classes.




\section{Conclusion}
\label{sec:conclusion}

In this paper, we proposed extreme amodal face detection, a new task that requires the model to detect and localize faces that are outside the image or truncated by the image frame. 
We construct the new EXAFace dataset for training and evaluating models on this task and propose a heatmap-based extreme amodal object detector with a novel selective coarse-to-fine decoder.
The results indicate that our approach outperforms other related methods, while requiring orders of magnitude less compute and memory.
This work points to the feasibility of efficiently inferring the presence of unseen objects, with possible applications in, for example, safer robot navigation, active surveillance, and realistic image expansion.

\newpage
\paragraph{Acknowledgments.}
Dr Campbell is the recipient of an Australian Research Council Discovery Early Career Award (project number DE250100542) funded by the Australian Government.

{
    \small
    \bibliographystyle{ieeenat_fullname}
    \bibliography{main}
}

\clearpage
\appendix
\section*{Supplementary Material}
\addcontentsline{toc}{section}{Supplementary Material}

\section{Complementary definitions and details}

\paragraph{Ground-truth heatmap generation.} Note that we generate the ground-truth heatmap from ground-truth bounding boxes with the same method as CenterNet~\cite {centernet}. 
In particular, we apply a Gaussian kernel on the center of bounding boxes, where the kernel size is calculated according to the box size.

\paragraph{Auxiliary task.} During training, our model will predict both faces and bodies, while in the evaluation, we only report the metrics regarding the faces.

\paragraph{Center sampling.} Recall that in \cref{eq:avg_pool} we define the average pooling positional encoding, now we introduce center sampling
\begin{equation}
    \text{cs}(s_i)(u,v) = \phi((\bar u, \bar v)),
\end{equation}
where it first average the coordinate of an $s_i \times s_i$ window and then encode it. When using center sampling, we replace it with $\text{avgpool}(\bm{p}, s_i)(u,v)$ with it in \eqref{eq:avg_pool}.
Since it discards the scale information, we adopt average pooling in our method.

\paragraph{Evaluation details.}
For the predicted bounding boxes, we apply a Non-Maximum-Suppression (NMS) IoU no more than 0.7, and retain the top-1000 predicted boxes based on confidence score. 
When evaluating the outpainting pipeline, we first apply the same NMS and top-1000 filter on the result of each image, then we aggregate all the remaining boxes and apply the NMS and filtering again. 
For the heatmap, we average the heatmaps over all images.

\section{Further discussion on the outpainting baseline}
\label{appendix:results}

\begin{figure*}[!t]
    \centering
    \includegraphics[width=\textwidth]{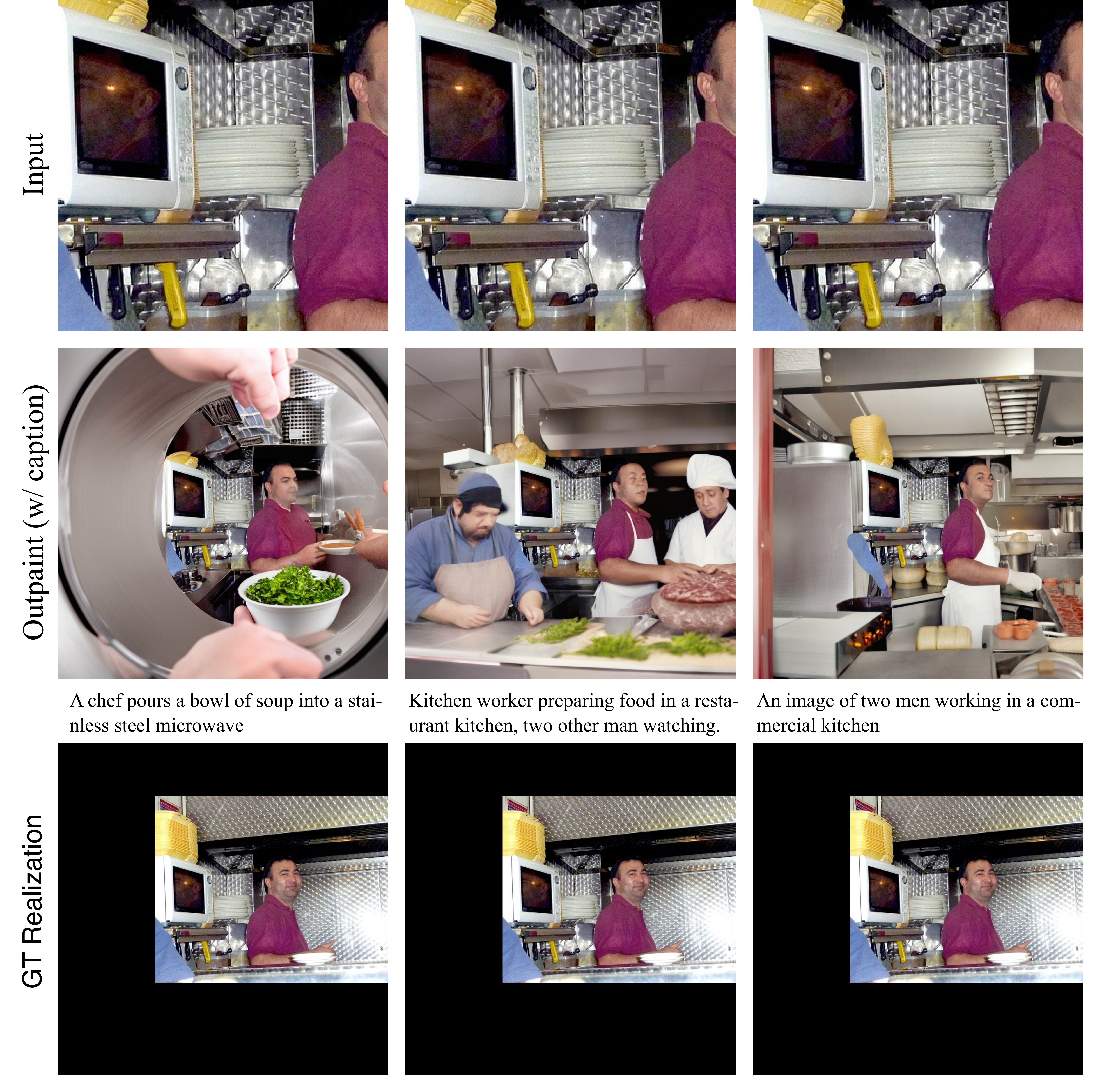}
    \caption{
    Outpainted example from SDXL~\cite{podell2023sdxl} + BLIP2~\cite{li2023blip}. 
    These three examples show that the outpainted example can be bottlenecked by any one component, and the randomness of the outpainted result. The middle example demonstrates that when both components collaborate well, the left and right example shows the bottleneck made by either VLM or the outpainting model.
    }
    \label{fig:ex_outpaint}
\end{figure*}

\cref{tab:ablation_outpaint} reports the performance of the outpainting pipeline with varying numbers of samples. Increasing the number of samples improves metrics for outside faces, but degrades CE and AP on truncated faces, revealing a trade-off inherent to this approach. A further limitation is that the pipeline is not end-to-end trainable, making each component a potential bottleneck (\cref{fig:ex_outpaint}). Moreover, even with strong generative models, accessing the ideal conditional distribution remains an open challenge.

\begin{table*}[!t]
{\footnotesize \begin{tabularx}{\textwidth}{@{}lCCCCCCCCCCCCCC@{}}
    \toprule
     Top-$\mu^2$                   & 
     AP$\uparrow$                  & AP$_\text{t}$$\uparrow$       & AP$_\text{o}$$\uparrow$       & 
     AP$_\text{o+}$$\uparrow$       &
     AP$_\text{o-}$$\uparrow$       &
     MAE$\downarrow$               &
     MAE$_\text{t}$$\downarrow$    &
     MAE$_\text{o}$$\downarrow$    &
     MAE$_\text{o+}$$\downarrow$    &
     MAE$_\text{o-}$$\downarrow$    &
     mIoU$\uparrow$                &
     Recall$\uparrow$              &
     CE$\downarrow$                &
     SE$\downarrow$\\
    \midrule
    15 & 21.37 & \underline{62.51} & 0.80 & 1.40 & 0.20 &
    23.99 & 2.33 & 37.08 & 5.64 & 48.53 &
    \underline{18.08} & 25.51 & \textbf{93.27} & \textbf{88.34} \\
    20 & 21.21 & 61.65 & \textbf{0.99} & \textbf{1.74} & \textbf{0.24} &
    18.59 & \underline{2.15} & 28.5 & 4.91 & 37.10 &
    17.56 & 26.35 & \underline{93.64} & 88.87  \\
    25 & \textbf{21.49} & \textbf{62.73} & 0.86 & 1.48 & \textbf{0.24} &
    19.69 & 2.19 & 30.28 & 4.82 & 39.55 & 17.84 & 26.64 & 93.78 & \underline{88.69}\\
    30 & 20.34 & 59.34 & 0.85 & 1.46 & \underline{0.23} &
    \textbf{16.31} & 2.19 & \textbf{24.94} & \underline{4.51} & \textbf{32.38} &
    \textbf{18.09} & \textbf{26.85} & 94.48 & 88.80 \\
    35 & 20.66 & 60.32 & 0.83 & 1.47 & 0.20 &
    \underline{17.29} & \textbf{2.07} & \underline{26.53} & \textbf{4.33} & \underline{34.61} &
    17.83 & 25.92 & 94.52 & 89.05 \\
    40 & \underline{21.38} & 62.35 & \underline{0.89} & \underline{1.56} & \underline{0.23} & 18.79 & 2.17 & 28.89 & 4.92 & 37.61 & 17.75 & 25.78 & 94.86 & 89.25 \\
    \bottomrule
\end{tabularx}}
\caption{Complete result of analysis on top-$\mu$ at scale $\cS=(2)$.}
\label{tab:ana_topu2}
\end{table*}

\begin{table*}[!t]
{\footnotesize \begin{tabularx}{\textwidth}{@{}lCCCCCCCCCCCCCC@{}}
    \toprule
     Scale                   & 
     AP$\uparrow$                  & AP$_\text{t}$$\uparrow$       & AP$_\text{o}$$\uparrow$       & 
     AP$_\text{o+}$$\uparrow$       &
     AP$_\text{o-}$$\uparrow$       &
     MAE$\downarrow$               &
     MAE$_\text{t}$$\downarrow$    &
     MAE$_\text{o}$$\downarrow$    &
     MAE$_\text{o+}$$\downarrow$    &
     MAE$_\text{o-}$$\downarrow$    &
     mIoU$\uparrow$                &
     Recall$\uparrow$              &
     CE$\downarrow$                &
     SE$\downarrow$\\
    \midrule
    $(1)$ & \underline{21.02} & \underline{61.28} & \textbf{0.89} & \textbf{1.59} & 0.19 &
    20.31 & 2.25 & 31.32 & 5.14 & 40.85 &
    \textbf{18.34} & 25.25 & \underline{96.36} & \underline{89.57} \\
    $(2)$ & 19.06 & 55.74 & 0.72 & 1.27 & 0.17 &
    21.28 & \underline{2.11} & 32.81 & 5.32 & 42.82 &
    17.60 & 23.67 & 96.62 & 90.34  \\
    $(4)$ & 19.93 & 58.62 & 0.58 & 1.03 & 0.12 &
    21.87 & 2.22 & 33.77 & 5.15 & 44.19 & 17.41 & 23.09 & 96.60 & 90.41 \\
    $(4,2)$ & 19.20 & 56.27 & 0.67 & 1.11 & \underline{0.22} &
    \textbf{13.64} & 2.27 & \textbf{20.68} & \underline{4.82} & \underline{39.55} &
    16.88 & \underline{25.34} & 98.66 & 94.06 \\
    $(2,1)$ & \textbf{21.49} & \textbf{62.73} & \underline{0.86} & \underline{1.48} & \textbf{0.24} &
    19.69 & 2.19 & 30.28 & \underline{4.82} & \underline{39.55} &
    17.84 & \underline{26.64} & \textbf{93.78} & \textbf{88.69} \\
    $(4,2,1)$ & 19.96 & 58.30 & 0.79 & 1.42 & 0.16 & \textbf{14.94} & \textbf{2.05} & \underline{22.86} & \textbf{4.63} & \textbf{29.49} & \underline{18.26} & 23.91 & 98.25 & 93.06 \\
    \bottomrule
\end{tabularx}}
\caption{Complete result of analysis on multiple-scale.}
\label{tab:ana_multiscale}
\end{table*}

\begin{table*}[!t]
{\footnotesize \begin{tabularx}{\textwidth}{@{}lCCCCCCCCCCCCCC@{}}
    \toprule
     Num of Samples                        & 
     AP$\uparrow$                  & AP$_\text{t}$$\uparrow$       & AP$_\text{o}$$\uparrow$       & 
     AP$_\text{o+}$$\uparrow$       &
     AP$_\text{o-}$$\uparrow$       &
     MAE$\downarrow$               &
     MAE$_\text{t}$$\downarrow$    &
     MAE$_\text{o}$$\downarrow$    &
     MAE$_\text{o+}$$\downarrow$    &
     MAE$_\text{o-}$$\downarrow$    &
     mIoU$\uparrow$                &
     Recall$\uparrow$              &
     CE$\downarrow$                &
     SE$\downarrow$\\
    \midrule
    1 & \textbf{9.07} & \textbf{24.01} & 1.59 & 2.01 & \textbf{1.17} &
    24.03 & 3.25 & 36.25 & 6.75 & 46.99 &
    18.41 & 24.91 & \textbf{93.68} & 92.56\\
    2 & \underline{7.75} & \underline{20.13} & 1.56 & \underline{2.23} & 0.89 &
    \underline{14.16} & 2.50 & \underline{20.95} & 4.57 & \underline{26.92} &
    19.98 & \textbf{25.27} & \underline{95.43} & 91.06\\
    5 & 5.89 & 15.02 & 1.32 & 1.86 & 0.78 &
    \textbf{13.75} & 2.17 & \textbf{20.45} & 3.71 & \textbf{26.54} &
    \underline{20.47} & \underline{25.15} & 96.18 & 90.39\\
    8 & 5.51 & 12.64 & \textbf{1.94} & 2.01 & \textbf{1.17} &
    \underline{14.16} & \underline{2.08} & 21.11 & \underline{3.58} & 27.50 &
    \textbf{20.53} & 25.07 & 96.35 & \underline{90.24}\\
    10 & 4.93 & 11.54 & \underline{1.62} & \textbf{2.47} & 0.76 & 14.69 & \underline{2.07} & 21.94 & \textbf{3.48} & 28.67 & \textbf{20.53} & 25.03 & 96.41 & \textbf{90.18} \\
    \bottomrule
\end{tabularx}}
\caption{Analysis of the number of outpainting samples.}
\label{tab:ablation_outpaint}
\end{table*}

\begin{figure*}[!t]
    \centering
    \includegraphics[width=\textwidth]{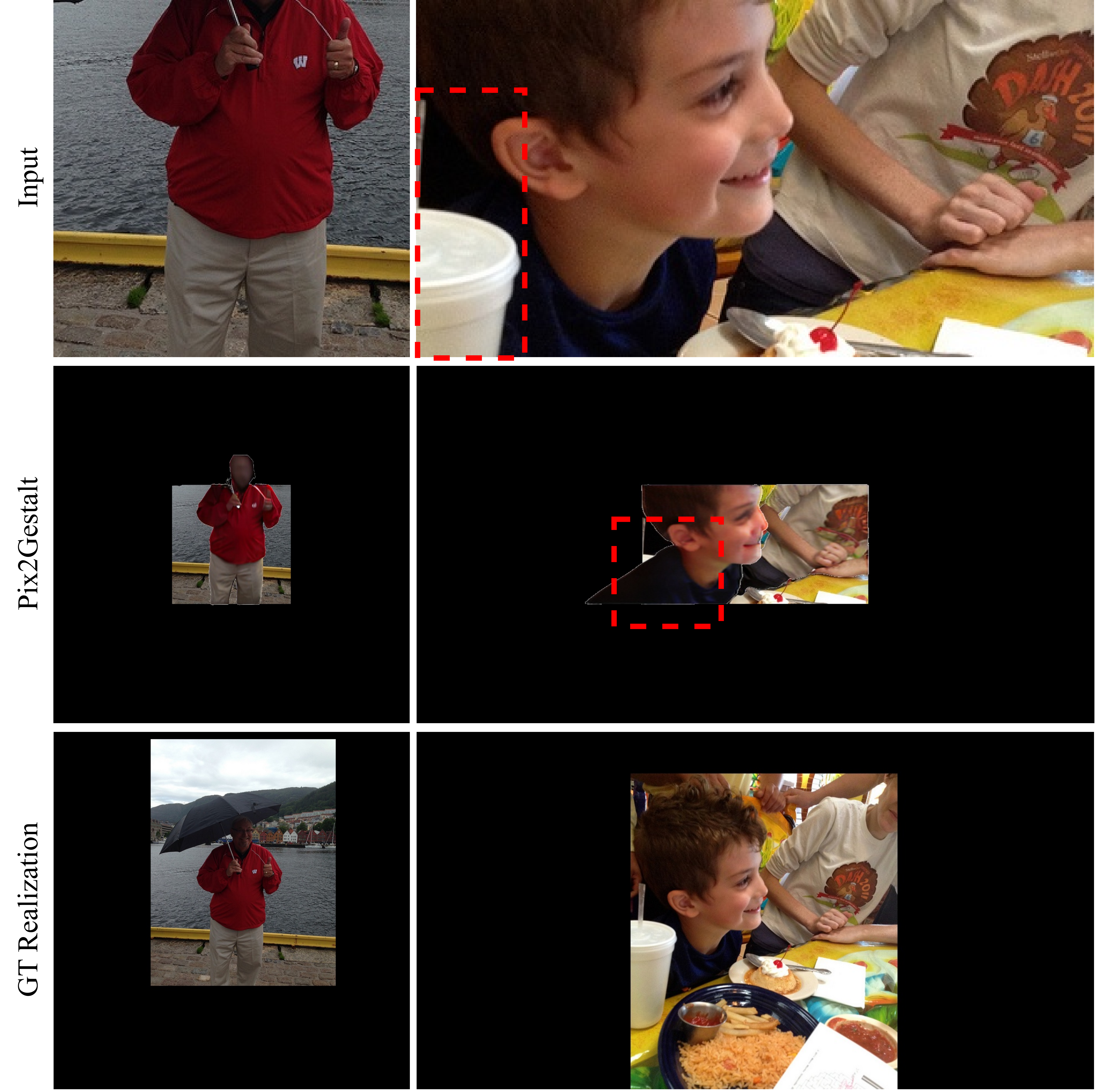}
    \caption{
    Completion examples with Pix2Gestalt~\cite{pix2gestalt}. 
    The first example shows that the model struggles to complete out-of-frame regions despite strong visual evidence, while the second demonstrates effective in-frame occluder removal. 
    Together, these cases highlight the distinction between in-frame completion and out-of-frame completion: 
    strong performance on the former does not necessarily transfer to the latter.
    }
    \label{fig:ex_pix2gestalt}
\end{figure*}

\section{Potential negative Societal impacts}

We also note the potential for more troubling applications (dual use). Successfully detecting objects like humans faces beyond what is directly observable could serve opposing ends. Instead of directing the camera to avoid that area, extreme amodal face detection could be used to pursue unseen-but-inferred objects. The existence of such applications does not negate the ethical case for extreme amodal face detection, though, which is based on its safety, privacy, and accessibility-enhancing potential.

\end{document}